\documentclass[10pt,twocolumn,letterpaper]{article}

\usepackage{wacv}
\usepackage{times}
\usepackage{epsfig}
\usepackage{graphicx}
\usepackage{amsmath}
\usepackage{amssymb}

\wacvfinalcopy % *** Uncomment this line for the final submission

\def\wacvPaperID{****} % *** Enter the wacv Paper ID here

\ifwacvfinal\fi
\begin{document}

%%%%%%%%% TITLE
\title{Object Specific Deep Learning Feature and Its Application to Face Detection}

% Authors at the same institution
%\author{First Author \hspace{2cm} Second Author \\
%Institution1\\
%{\tt\small firstauthor@i1.org}
%}
% Authors at different institutions
\author{Xianxu Hou \\
University of Nottingham, Ningbo, China\\
{\tt\small xianxu.hou@nottingham.edu.cn}
\and
Ke Sun \\
University of Nottingham, Ningbo, China\\
{\tt\small ke.sun@nottingham.edu.cn}
\and
Linlin Shen \\
Shenzhen University, Shenzhen, China\\
{\tt\small llshen@szu.edu.cn}
\and
Guoping Qiu \\
University of Nottingham, Ningbo, China\\
{\tt\small guoping.qiu@nottingham.edu.cn}
}

\maketitle
\ifwacvfinal\thispagestyle{empty}\fi

%%%%%%%%% ABSTRACT
\begin{abstract}
We present a method for discovering and exploiting object specific deep learning features and use face detection as a case study. Motivated by the observation that certain convolutional channels of a Convolutional Neural Network (CNN) exhibit object specific responses, we seek to discover and exploit the convolutional channels of a CNN in which neurons are activated by the presence of specific objects in the input image. A method for explicitly fine-tuning a pre-trained CNN to induce an object specific channel (OSC) and systematically identifying it for the human face object has been developed. Based on the basic OSC features, we introduce a multi-resolution approach to constructing robust face heatmaps for fast face detection in unconstrained settings. We show that multi-resolution OSC can be used to develop state of the art face detectors which have the advantage of being simple and compact.
\end{abstract}

%%%%%%%%% BODY TEXT
\section{Introduction}
\label{sec:intro}
%Face Detection is a well-studied problem and frontal faces can be easily detected in real time by modern detectors such as the seminal work of Viola and Jones (VJ) \cite{viola2001rapid}. However, VJ detector doesn't work very well in unconstrained settings due to variations in poses, illuminations and occlusions. Robust face detection in unconstrained settings remains to be a challenging problem.

Convolutional Neural Networks (CNNs) have been demonstrated to have state-of-the-art performances in many computer vision tasks such as image classification \cite{krizhevsky2012imagenet,simonyan2014very}, detection \cite{girshick2014rich,sermanet2013overfeat}, retrieval \cite{babenko2014neural} and captioning \cite{karpathy2015deep}. % based on millions of parameters and massive datasets. 
Several methods such as Deepvis \cite{yosinski2015understanding} and deconvolutional technique \cite{zeiler2014visualizing} have been developed to help get a better understanding of how these models work and it has been shown that CNNs have the ability to learn powerful and interpretable features. 

\begin{figure*}
\begin{tabular}{ccc}
\rule{0pt}{1ex}\hspace{2.24mm}\includegraphics[width=16cm]{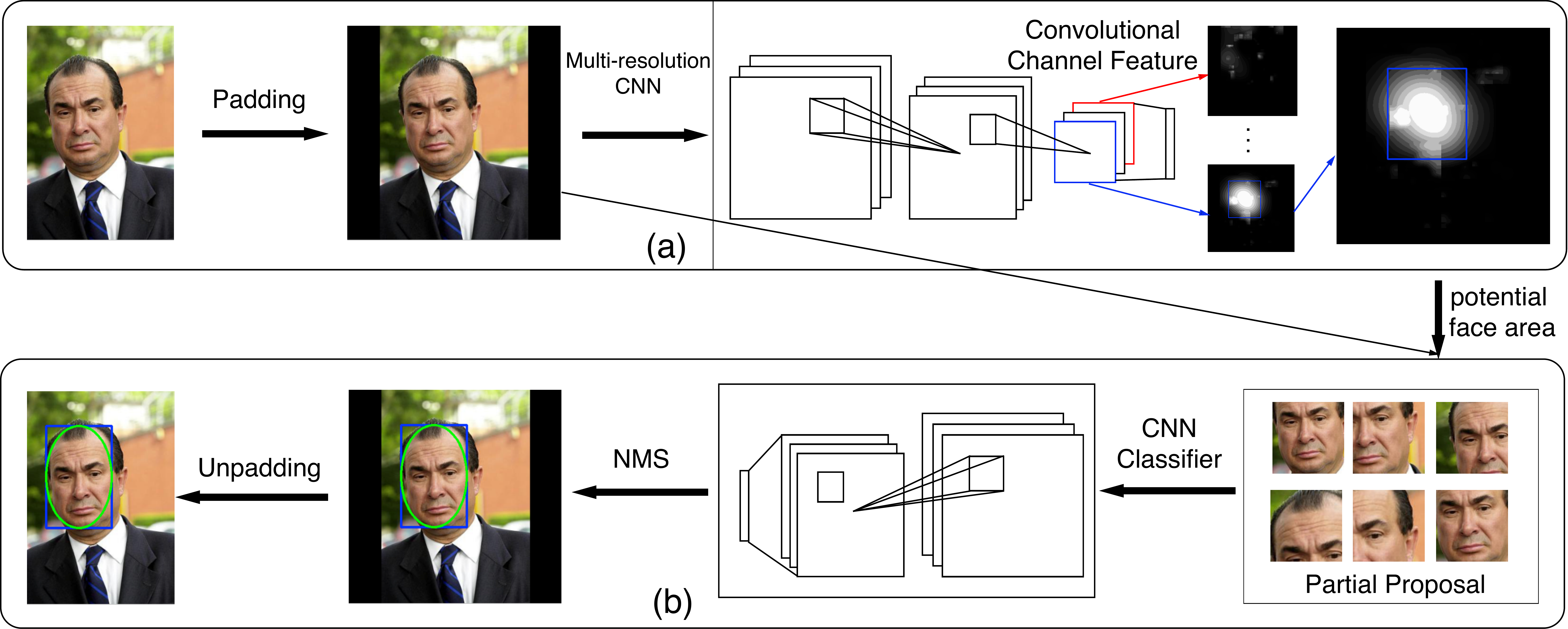}\\[-0.1pt]
\end{tabular}
\caption{(a) An input image is first processed by a CNN, its face specific convolutional channels are identified and a face response heatmap is generated from multi-resolution face specific convolutional channel features. (b) Face proposals are generated based on the heatmap and processed by a binary CNN classifier. Finally detected faces are combined through a Non-Maximum Suppression (NMS) algorithm.}
\label{fig:overview}
\end{figure*}

In this paper, we present a method for discovering and exploiting class specific deep learning features and use face detection as a case study. A key motivation of this paper is based on the observation that certain convolutional channels of CNNs exhibit object specific responses \cite{yosinski2015understanding}. An object specific channel (OSC) is a convolutional feature map at a hidden layer of a CNN in which neurons are strongly activated by the presence of a certain class of objects at the neurons' corresponding regions in the input image. An example is shown in Figure \ref{fig:overview} where the last image at the top row is a face specific OSC where spatial locations corresponding to the face regions have strong responses (white pixels) while areas corresponding to non-face regions have weak responses (black pixels). If such an OSC can be reliably identified, then it can be exploited for various tasks including object detection. Do such channels exist for a given class of objects? If so, how can we systematically identify such channels? If not, can we tune a pre-trained CNN to have such channels?   

A method for explicitly fine-tuning a pre-trained CNN to induce an OSC and systematically identifying it for the human face object has been developed. Based on the basic OSC features, we introduce a multi-resolution approach to constructing robust face heatmaps for fast face detection in unconstrained settings. Although we use face as a specific case study, our method could be easily extended to other classes of objects to turn the convolutional feature maps of a CNN into object specific feature channels for object detection and other computer vision tasks.
 
%We propose a two-stage model for face detection in unconstrained settings. In the first step, features in a single channel in one convolutional layer are used to locate candidate facial regions by a score mechanism, and then a CNN binary classifier is used to refine face proposals and eliminate noises. Concretely our contributions are three-fold:
%\begin{itemize}
%  \item We propose to use a multi-resolution approach to construct face response heatmap using single convolutional channel features. The heatmap is like a contour map corresponding to potential face area in the input image. We can directly use the heatmap for real time face detection, or improve accuracy by adding a binary CNN classifier to refine the detected results.
%  \item Our simple model shows robustness to various scales, occlusions, poses and illuminations. It doesn't need extra image information such as landmarks and pose annotations, or extra model components such as bounding box regression.
%  \item We show that manually designed data augmentation is effective to tackle different facial variations by making full use of the rich representative capacity of CNNs.
%\end{itemize}

\section{Related Work}
Face detection models in the literature can be divided into four categories: Cascade-based model, Deformable Part Models (DPM)-based model, Exemplar-based model and Neural-Networks-based model. The most famous cascade-base model is the VJ detector \cite{viola2001rapid} based on Haar-like features, which have demonstrated excellent performance in frontal face detection. However Harr-like features have limited representation ability to deal with variational settings. Some works try to improve VJ-detector via using more complicated features such as SURF \cite{zhu2006fast}, HoG \cite{li2013learning} and polygonal Haar-like features \cite{pham2010fast}. Aggregate channel features \cite{yang2014aggregate} are also introduced for solving multi-view face detection problems.

Another category is DPM-base model \cite{felzenszwalb2010object}, which treats face as a collection of small parts. DPM-base model can benefit from the fact that different facial parts independently have lower visual variations, so it is reasonable to build robust detectors by combining different models trained for individual parts. For example, Part-based structural models \cite{yan2014face,zhu2012face,yan2014fastest} have achieved success in face detection and a vanilla DPM can achieve top performance over the more sophisticated DPM variants \cite{mathias2014face}.

Exemplar-based detectors \cite{shen2013detecting,li2014efficient,kumar2015visual} try to bring image retrieval techniques into face detection to avoid explicitly modelling different face variations in unconstrained settings. Specifically, each exemplar casts a vote following the Bag-of-Words (BOW) \cite{shekhar2012word} retrieval framework to get a voting map and uses generalized Hough Voting \cite{leibe2004combined} to locate the faces in the input image. As a result, faces can be effectively detected in many challenging settings. However, a considerable amount of exemplars are required to cover all kinds of variations.

Neural-Networks-based detectors are usually based on deep convolutional neural networks. Faceness \cite{yang2015facial} tries to find faces through scoring facial parts responses by their spatial structure and arrangement, and different facial parts correspond to different CNNs. A two-stage approach is also proposed by combining multi-patch deep CNNs and deep metric learning \cite{liu2015targeting}. The CCF detector \cite{yang2015convolutional} uses an integrated method called Convolutional Channel Features, transferring low-level features extracted from pre-trained CNN models to a boosting forest model. Cascade architectures based on CNNs \cite{li2015convolutional} have been also designed to help reject background regions at low resolution, and select face area carefully at high resolution. The DDFD detector \cite{farfade2015multi} uses a single model based on deep convolutional neural networks for multi-view face detection, and points out that CNNs can benefit from better sampling and more sophisticated data augmentation techniques.

We try to directly use object specific channel to produce face response heatmap, which can be used to quickly locate potential face area. The heatmap is similar to the voting map in exemplar-based approach \cite{li2014efficient,kumar2015visual}, but the difference is that the voting map is produced by the Bag-of-Words (BOW) \cite{shekhar2012word} retrieval framework, while our heatmap is directly extracted from a convolutional channel.

\begin{figure*}
\begin{tabular}{ccc}
\rule{0pt}{1ex}\hspace{2.24mm}\includegraphics[width=16cm]{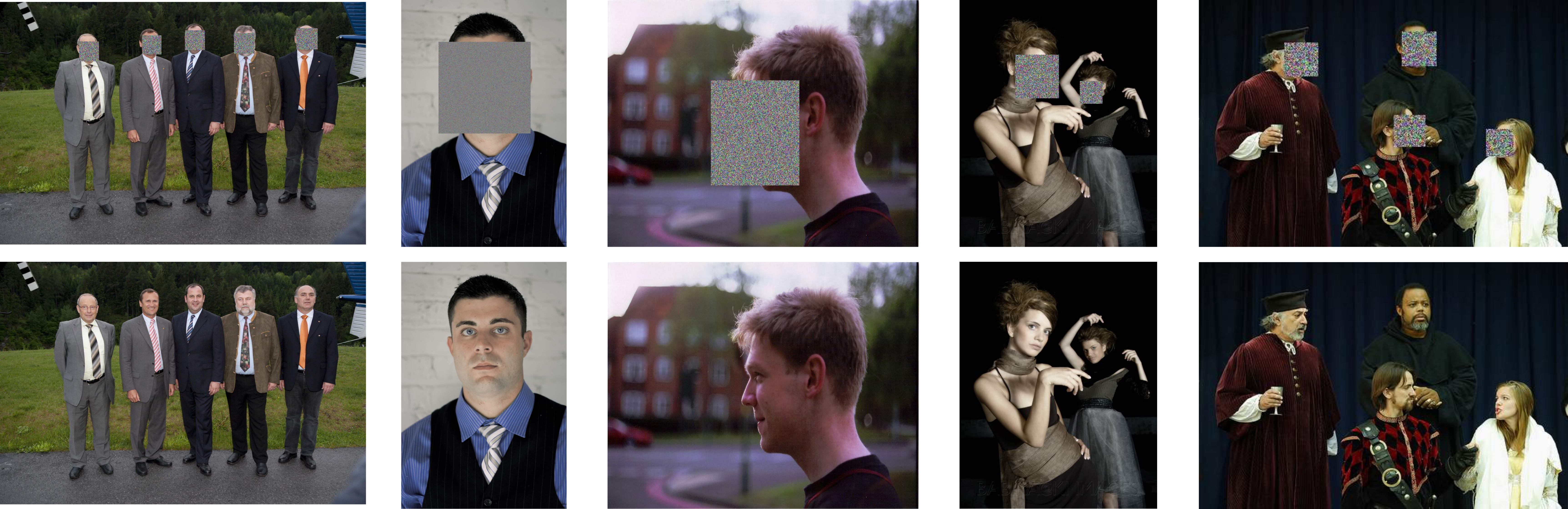}\\[-0.1pt]
\end{tabular}
\caption{Examples of masked face images (first row) and original images (second row) from AFLW \cite{kostinger2011annotated} for fine-tuning.}
\label{fig:maskedFace}
\end{figure*}

%\section{Our Approach:Tuning Convolution Features for Face Detection}
\section{Object Specific Convolutional Features for Face Detection}
\subsection {Overview}
%This section we introduce Our ContourFace detector. 
Our goal is to discover and exploit face specific convolutional channel features to help locate the face areas quickly for further processing. 
Our face specific feature discovery and detection architecture contains two stages as shown in Figure \ref{fig:overview}: In the first stage, an image is fed to a CNN. face specific convolutional channels (OSCs) are identified. Multi-resolution features are extracted from the OSCs to form the channel's face heatmap. In the second stage, a set of face candidate windows can be quickly identified based on the OSC heatmap. All candidates are then processed by a CNN based binary classifier. Finally all face windows are merged using Non-Maximum Suppression (NMS) \cite{neubeck2006efficient} to obtain the final detection results.

\subsection {face Specific CNN Feature Extraction}

{\bf Training Data Preparation and CNN Fine-tuning.} We start with the pre-trained "AlexNet" \cite{krizhevsky2012imagenet} provided by the open source Caffe Library \cite{jia2014caffe}. In order to adapt the CNN model to our face detection problem, we change the last classification layer from ImageNet-specific 1000 classes to 2 classes, which represent images with faces and images with masked faces (Figure \ref{fig:maskedFace}) respectively. Specifically, let Conv be a convolutional layer, LRN a local response normalization layer, P a max pooling layer and F a fully connected layer, the architecture can be described as Conv1(55x55x96) - LRN - P - Conv2(27x27x256) - LRN - P - Conv3(13x13x384) - Conv4(13x13x384) - Conv5(13x13x256) - P - F(4096) - F(4096) - F(2), where the numbers inside the brackets are the numbers of neurons and their topology).

The authors of \cite{yosinski2015understanding,zeiler2014visualizing} tried to gain insight into the operation of the CNN models via visualizing the features and filters of their hidden units. The authors of \cite{zeiler2014visualizing} have shown that the CNN model is highly sensitive to local structure of the input image and that particular regions of an image are responsible for firing of specific neural units. \cite{yosinski2015understanding} discovers that there exists many invariant detectors for faces, shoulders, text, etc. in the $5^{th}$ convolutional layer. It shows that CNNs can learn partial information even though no explicitly labelled faces or texts exist in the training datasets. 

If we can be certain that specific objects will fire specific hidden neurons in a CNN then it will be very useful because we can infer the objects in the input from the response of the hidden neurons. Based on above observations, we believe it is possible to fine-tune a CNN such that particular neurons will respond to specific objects if suitably prepared object specific training examples were used. To verify this idea, we use Annotated Facial Landmarks in the Wild (AFLW) database \cite{kostinger2011annotated} as training dataset to fine-tune the AlexNet. AFLW database contains 25,993 faces in 21,997 real world images collected from flickr with a range of diversity and variation in poses, ages and illuminations. Unlike most common methods that crop the ground-truth face area as positive samples and non-face area as negative samples, we use the original images as positive samples. For negative samples, we also use the same images, but mask the facial part with random noises (R, G, B value are randomly generated from 0-255). Some examples of these positive and negative samples are shown in Figure \ref{fig:maskedFace}. The idea is that if the CNN model can discriminate the masked image from the original image, it will be forced to use the partial information from face area for classification. What we desire is that through re-enforcing the CNN to discriminate images with faces and images without faces, then the network will organise itself such that certain units will be responsible for representing faces, thus enabling the extraction of face specific channels for various post processing. %Finally we hope that we could get better facial related convolutioanl channel features.

\begin{figure*}[!htb]
\centering
\begin{minipage}{.55\textwidth}
  \centering
  \includegraphics[width=6cm]{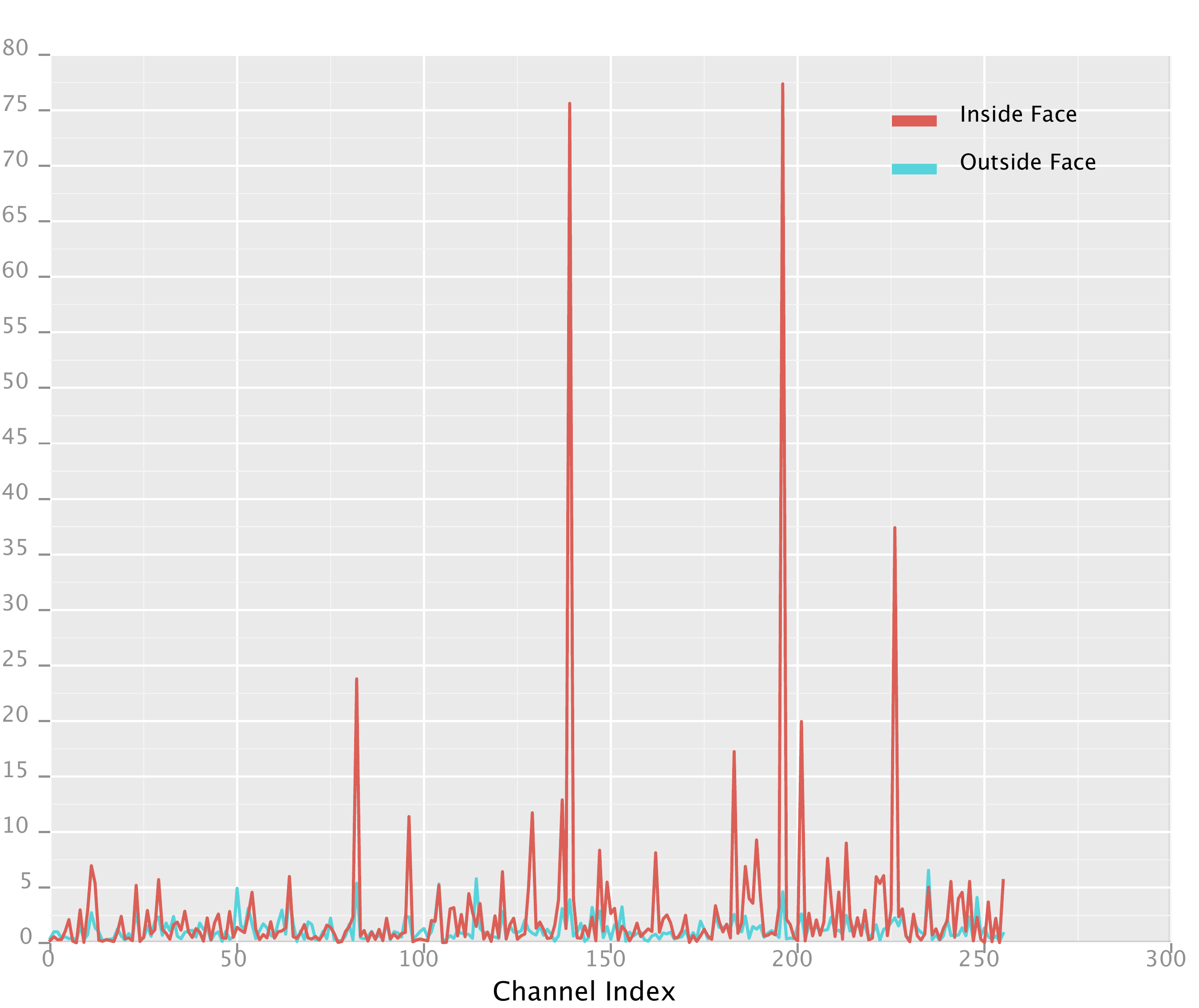}
  \caption{Face-scores of inside and outside face area for each channel in $conv5$ layer, averaged over 1,000 images randomly chosen from ALFW}
  \label{fig:face_score}
\end{minipage}%
\hfill
\begin{minipage}{.4\textwidth}
  \centering
  \includegraphics[width=5cm]{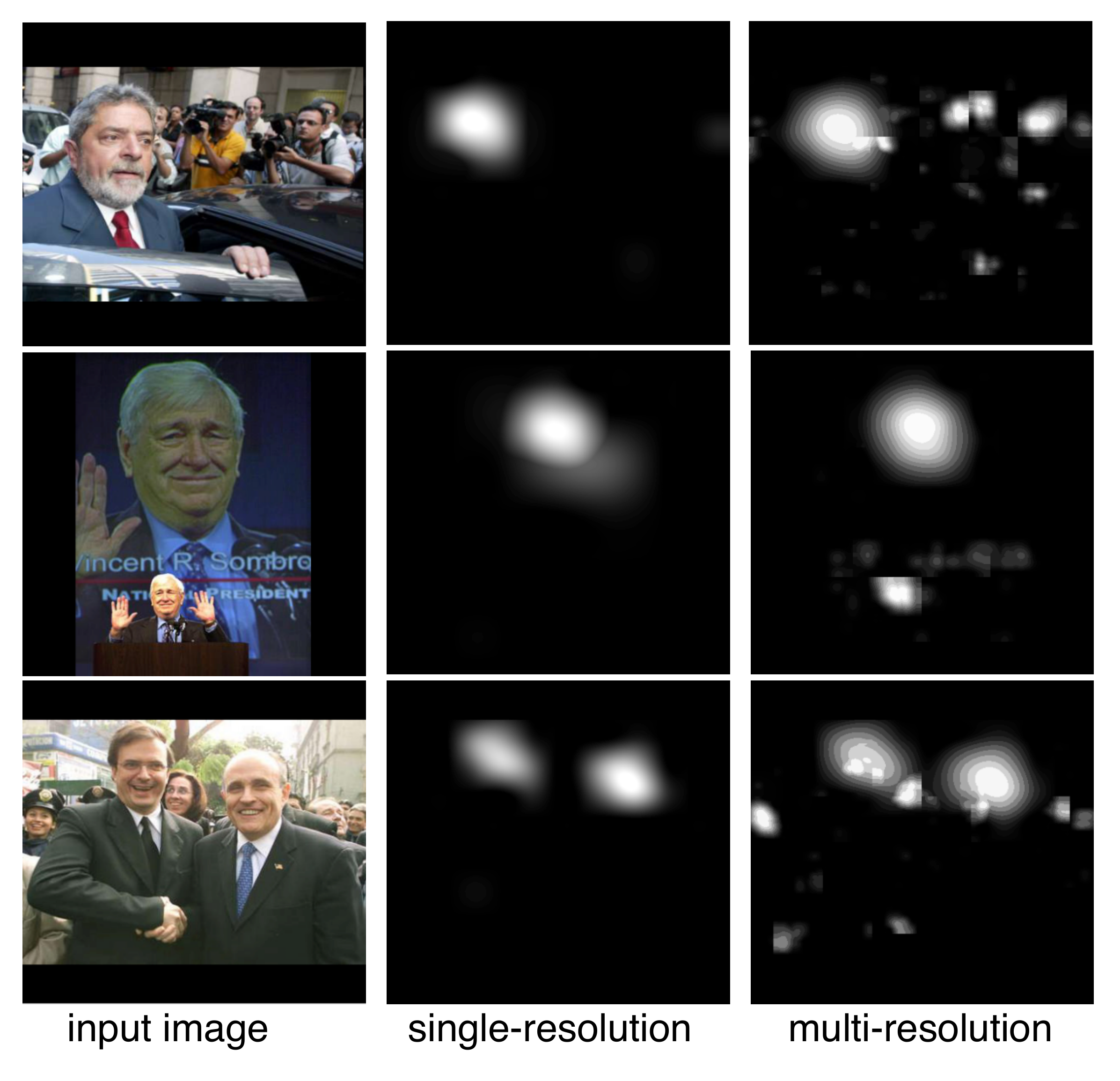}
  \caption{Examples for single-resolution and mutli-resolution feature extraction}
  \label{fig:multi_resolution}
\end{minipage}
\end{figure*}

{\bf Face Specific Convolutional Feature Extraction.} In order to quantify how well each convolutional channel responds to faces after fine-tuning, we've studied the $5^{th}$ convolutional layer which has a 13 x 13 x 256 topology and can be visualized as 256 different 2-D heatmaps (13 x 13). We resize every heatmap from shape (13 x 13) to (227 x 227), the same size as the input image, using a bicubic interpolation. We calculate the average intensity value of the heatmap both inside and outside the face areas respectively, which are denoted as "face-score", i.e., $\frac{1}{wh}\sum_{i=x}^{x+w} \sum_{j=y}^{y+h} I(i, j)$, for a given bounding box, where (x, y) is the top left coordinate, (w, h) are the width and height, and I(i, j) is the intensity value at point (i, j). The resized heatmap has the same shape as the input image, and the face area in the heatmap can be located by directly using the ground-truth face annotations. To calculate the face-score in the face areas, all the intensity of pixels inside the face area are added up and divided by the number of pixels in that face area. Similarly, the face-score outside the face area can be calculated easily. We then use 1,000 images randomly chosen from AFLW \cite{kostinger2011annotated} to calculate the face-scores inside and outside face areas of all the channels in the $5^{th}$ convolutional layer of the fine-tuned model. The final value of face-score are the average value across all the 1,000 images. The results are shown in Figure \ref{fig:face_score}. We can see that the $196^{th}$ channel has highest face-score inside the face areas, followed by the $139^{th}$ channel. The value of face-score outside face area is small in all channels. 

% Additionally face-scores of $139^{th}$ and $196^{th}$ inside face area are increased by more than 30\% after fine-tuning, while the face-scores outside face area remain small. 

This shows that there do exist face specific channels where specific neurons will respond to specific objects. What is also interesting is that the topology of the convolutional layer has a direct correspondence with the input image where objects (faces in this case) fires neurons at the spatial positions corresponding to face regions in the input image.

%Through fine-tuning the CNN with object specific training samples, object specific channels can be further strengthened, in this case, the face-scores inside the face regions have been boosted by over 30\% demonstrating the usefulness of our approach to preparing object specific training samples for extracting object specific convolutional features.  %It means that the response area in the heatmap can have a better chance to represent the ground-truth face area in the input image, and CNN could be forced to learn partial information from face area by fine-tuning with masked face images. In general, the channels with high face-score inside face area and low value outside the face area could be considered responding to the facial part of input image. Therefore, we use $196^{th}$ channel as our convolutional channel feature. 

% \begin{figure*}[!htb]
% \begin{tabular}{ccc}
% \rule{0pt}{1ex}\hspace{2.24mm}\includegraphics[width=16cm]{images/face_score_hist.pdf}\\[-0.1pt]
% \end{tabular}
% \caption{Face-scores inside and outside the face areas of the $5^{th}$ convolutional layer, averaged over 1,000 images randomly chosen from ALFW}% for pretrained and fine-tuned CNN model.}
% \label{fig:face_score}
% \end{figure*}

% \begin{figure*}
% \begin{tabular}{ccc}
% \rule{0pt}{1ex}\hspace{2.24mm}\includegraphics[width=16cm]{images/multi_resolution.pdf}\\[-0.1pt]
% \end{tabular}
% \caption{Examples for mutli-resolution feature extraction}
% \label{fig:multi_resolution}
% \end{figure*}

{\bf Mutli-resolution Features.} From the above, we could compute contours of the heatmap ($conv5_{196}$) based on a given threshold, and then generate the bounding box of that contour. As a result, the bounding box could be used to represent the face position. However, as shown in Figure \ref{fig:multi_resolution}, there are two problems if we want to use the heatmap directly for face detection. One is that the heatmap cannot capture small faces, the other is that the strong firing neurons in the heatmap tend to shift to the edge of the image if the face is not in the center of an image. We have developed a multi-resolution approach to solving these problems: The input image is divided into small sub-images by a sliding window at multiple resolutions. We specify the stride, with which the sliding window is moved each time, to be half of sub-image size. Thus adjacent sub-images overlap each other. All sub-images and the original image are then resized and fed to the fine-tuned CNN to extract the channel heatmaps ($conv5_{196}$), which are then merged to a single heatmap by selecting the maximum intensity value of individual heatmap at each pixel position, i.e., $I_{i, j} = \max\limits_{1\leq l\leq n} I^{l}_{i, j}$, where $I_{i, j}$ is the intensity value of the merged heatmap at point (i, j), and $I^{l}_{i, j}$ is the intensity value of the $l^{th}$ sub-image's heatmap at point (i, j). As a result (see Figure \ref{fig:multi_resolution}), the merged heatmap is able to capture small faces and locate faces in the input image more precisely. This is because small faces in the original scale become "larger" in the sub-image scale, which can be captured by the fine-tuned CNN. Furthermore, after performing color quantization \cite{verevka1996local} over the merged heatmap to reduce the number of distinct grey colors, the heatmap is like a contour map (see Figure \ref{fig:multi_resolution} and Figure \ref{fig:detected_examples}) corresponding to face area in the input image. The intensity value of pixels is highest around the eyes and noses area in the input image, and gradually drops off in other areas.

% Furthermore, we perform color quantization \cite{verevka1996local} over the merged heatmap to reduce the number of distinct grey colors. We first apply K-means clustering algorithm to reduce the grey colors from 255 to K (parameter used in K-means, 40 in our case), and then allocate the quantized color to the centroid which the input pixel is closest to. Interestingly, the heatmap is like a contour map (see Figure \ref{fig:multi_resolution} and Figure \ref{fig:detected_examples}) corresponding to face area in the input image after color quantization. The intensity value of pixels is highest around the eyes and noses area in the input image, and gradually drops off in other areas.

% \begin{figure*}
% \begin{tabular}{ccc}
% \rule{0pt}{1ex}\hspace{2.24mm}\includegraphics[width=16cm]{images/multi_resolution.pdf}\\[-0.1pt]
% \end{tabular}
% \caption{Examples for mutli-resolution feature extraction}
% \label{fig:multi_resolution}
% \end{figure*}

\begin{figure*}
\begin{tabular}{ccc}
\rule{0pt}{1ex}\hspace{2.24mm}\includegraphics[width=16cm]{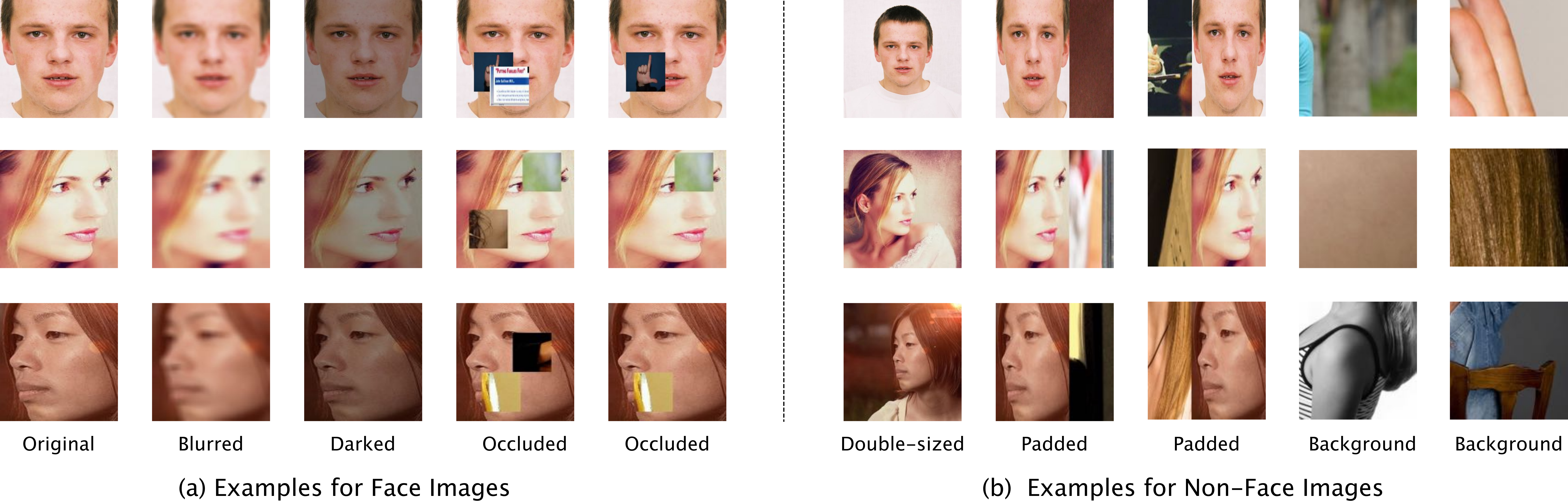}\\[-0.1pt]
\end{tabular}
\caption{Example face (a) and non-face (b) images from AFLW for face classifier training.}
\label{fig:training_example}
\end{figure*}

\subsection {Face Proposals and Detection}

The second stage of our method (Figure \ref{fig:overview}(b)) locates faces more precisely and separates overlapped faces which cannot be distinguished in the heatmap. We first use face-score defined above (average intensity of a specific area in a heatmap) to quickly select face proposals. Then another binary CNN classier is trained to discriminate face from non-face images. In the end, Non-Maximum Suppression is used to merge multiple face windows for final detection.

{\bf Face Proposals by Face-Score.} Instead of using category-independent region proposals method like selective search \cite{uijlings2013selective} used in R-CNN \cite{girshick2014rich} for general object detection, we simply use a multi-scale sliding window to select potential face windows based on face-score. Specifically, while scanning the input image, the face-score of the sliding window is calculated at the same position in the corresponding heatmap. The sliding windows are selected as face region proposals if the face-scores exceed a given threshold (80 in our case). This approach can reject non-face regions effectively based on the above observation that the pixel intensity values around the face regions are higher than non-face regions. All potential face proposals of one image are collected as a batch to feed to a binary CNN classifier described below.

{\bf Candidate Face Window Selection by CNN.} Due to a range of visual variations such as poses, expressions, lightings and occlusions, a robust face classifier is trained using augmented data. The fine-tuned CNN model used to extract face response heatmap is used as pre-trained model, and then fine-tuned again with face images and non-face images. All the face images are cropped from AFLW dataset \cite{kostinger2011annotated} by the ground-truth bounding box, and augmented by making them darkened, blurred and occluded (Figure \ref{fig:training_example}(a)). Non-face images are collected in the following ways (Figure \ref{fig:training_example}(b)): (1) background images randomly cropped from AFLW images with a given Intersection-over-Union (IoU) ratio (0, 0.1 and 0.2) to a ground-truth face; (2) cropped by double-sized ground-truth faces; (3) face images padded with non-face images. Each condidate face window has a classification score after being processed by the fine-tuned binary classifier. Finally all detected face windows are sorted from highest to lowest based on classification score, and then a non-maximum suppression (NMS) is applied to the detected windows to reject the window if it has an Intersection-over-Union (IoU) ratio bigger than a given threshold.

\section{Experiments}
As described above, the AFLW dataset is used to train our model, and then we use PASCAL Face \cite{yan2014face} and FDDB dataset \cite{fddbTech} to evaluate our face detector. PASCAL Face dataset is a widely used face detection benchmark, containing 851 images and 1,341 annotated faces. FDDB dataset is a larger face detection benchmark, consisting of 5,171 annotated faces in 2845 images. It contains a wide range of difficulties including occlusions, various poses and out-of-focus faces.

\begin{figure*}
\begin{tabular}{ccc}
\rule{0pt}{1ex}\hspace{2.24mm}\includegraphics[width=16cm]{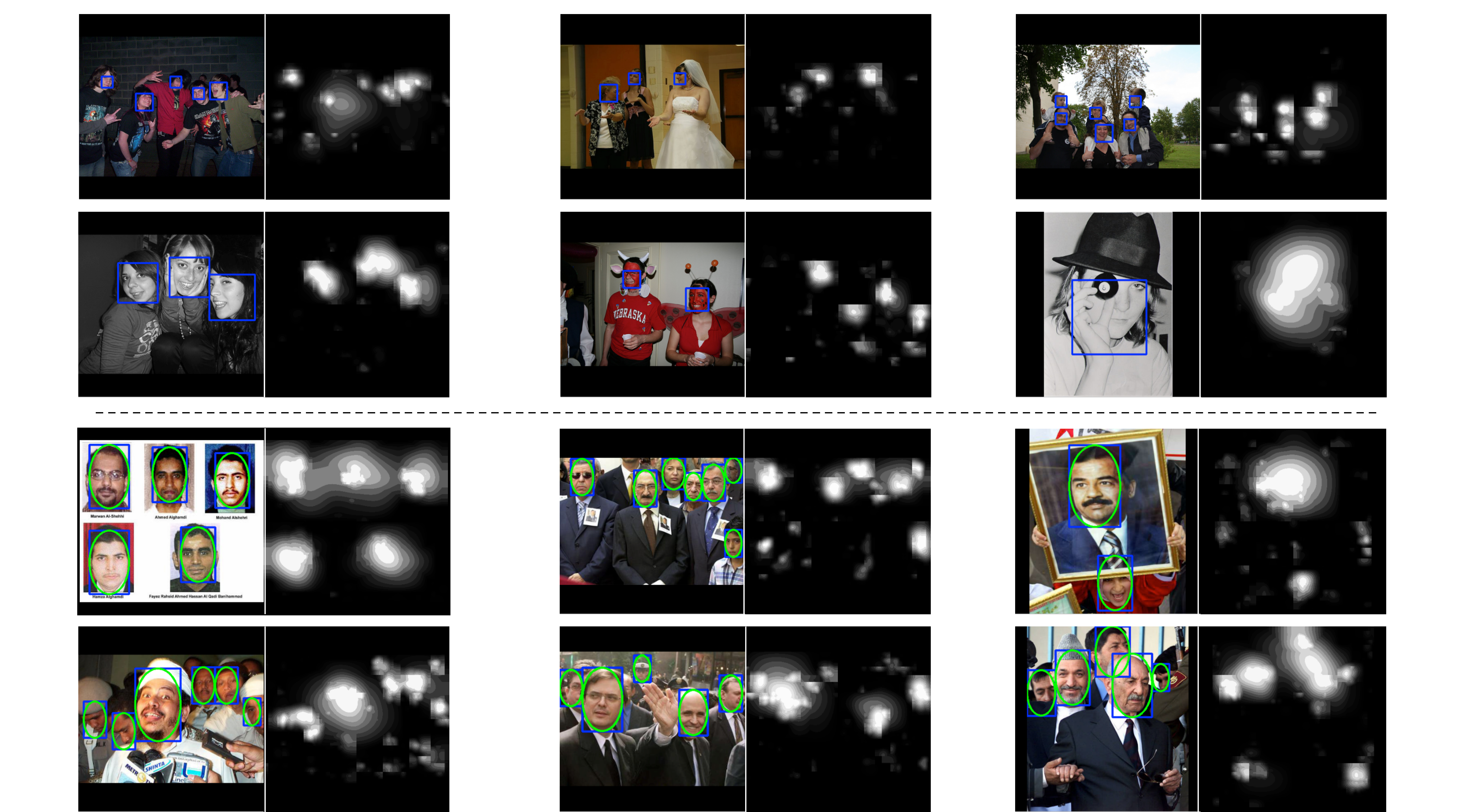}\\[-0.1pt]
\end{tabular}
\caption{Qualitative face detection results and face response heatmap by our detector on PASCAL face (top two rows) and FDDB dataset (bottom two rows).}
\label{fig:detected_examples}
\end{figure*}

One problem in the evaluation of face detection is the different annotations between training datasets and testing datasets such as policies for what constitutes a face, size of annotation boxes and minimum/maximum face size. In order to solve this problem, some works \cite{li2015convolutional,mathias2014face, yang2014aggregate} try to manually adjust the annotations to get better results. In our work, we use the original and adjusted annotation \cite{mathias2014face} of PASCAL Face dataset with the toolbox provided by \cite{mathias2014face}. FDDB dataset is evaluated with the original elliptical annotations using two evaluation protocol provided by the report \cite{fddbTech}: the continuous score and discontinuous score. In order to better fit the elliptical annotations that cover the whole faces, we extend our detected square boxes vertically by 40\% to upright rectangles. We also fit the largest upright ellipses for the extended rectangles as elliptical outputs for evaluation (Figure \ref{fig:detected_examples}).

\begin{figure*}[!htb]
\begin{tabular}{ccc}
\rule{0pt}{1ex}\hspace{2.24mm}\includegraphics[width=16cm]{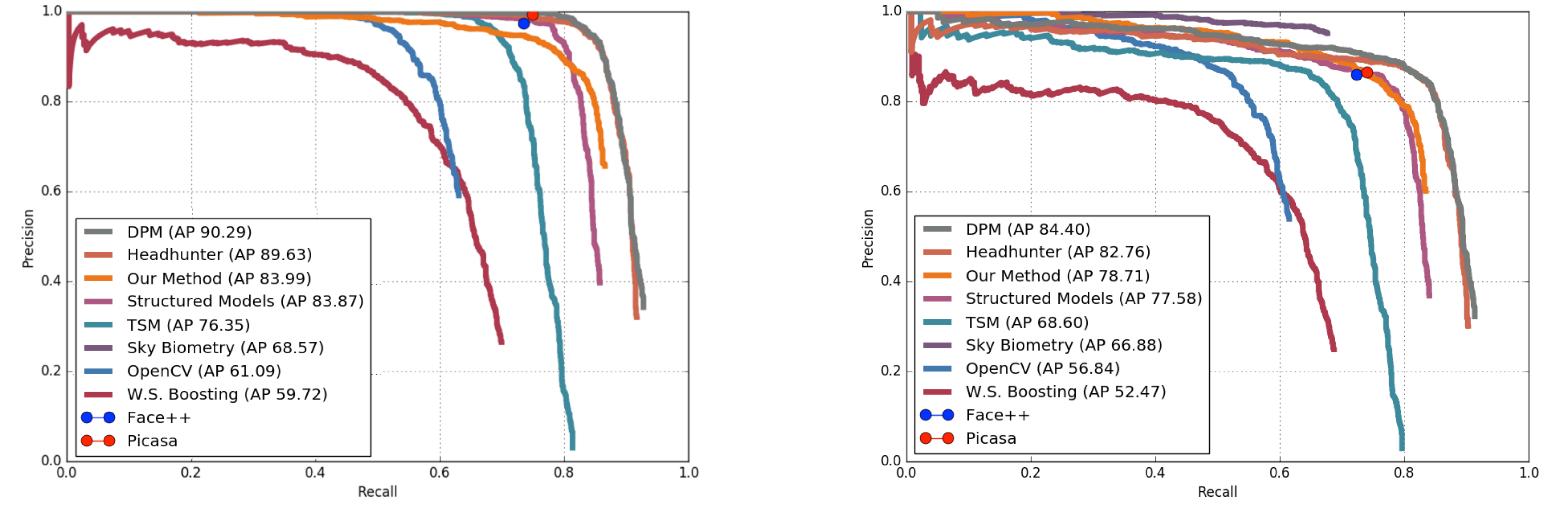}\\[-0.1pt]
\end{tabular}
\caption{Performance comparison on PASCAL Face dataset}
\label{fig:result_pascal}
\end{figure*}

\begin{figure*}
\begin{tabular}{ccc}
\rule{0pt}{1ex}\hspace{2.24mm}\includegraphics[width=16cm]{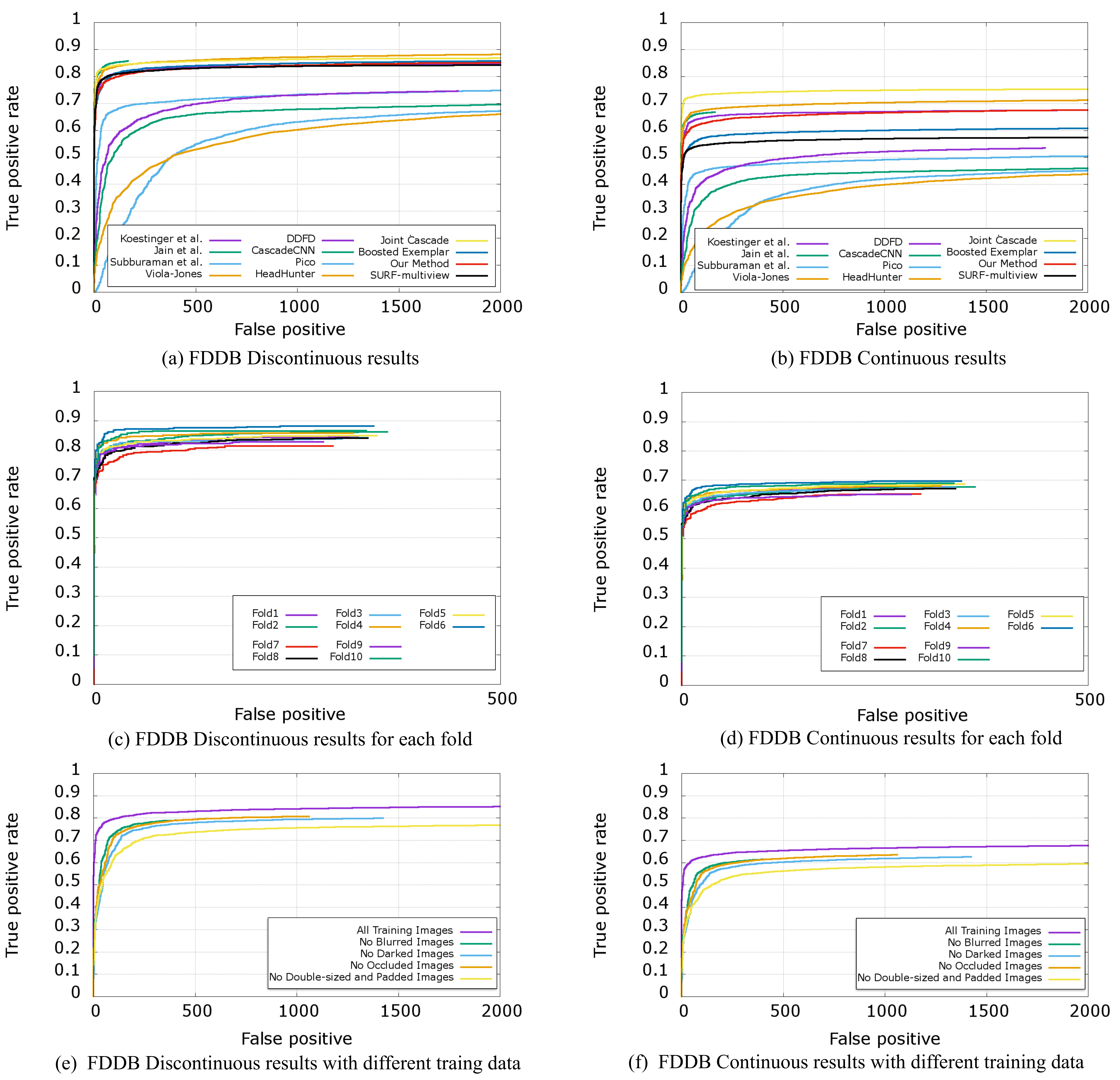}\\[-0.1pt]
\end{tabular}
\caption{Performance comparison on FDDB dataset with discrete and continuous protocols.}
\label{fig:result_fddb}
\end{figure*}

We report the average precision on PASCAL Face dataset (Figure \ref{fig:result_pascal}), continuous and discontinuous ROC for all the 10 folds (Figure \ref{fig:result_fddb} (a) and (b)) and individual ROC (Figure \ref{fig:result_fddb} (c) and (d)) for each fold on FDDB dataset. By comparing with other state-of-the-art face detectors \cite{farfade2015multi, li2015convolutional, markuvs2013method, chen2014joint, li2014efficient, li2013learning, kostinger2012robust, jain2011online, subburaman2010fast}, our method can achieve similar results both on PASCAL faces \cite{yan2014face} and FDDB \cite{fddbTech} datasets while having minimal complexity. 

Compared to many methods in the literature, our method is much simpler: ground-truth bounding boxes are the only information needed to train our model, and one single model can capture all the facial variations based on the carefully designed data augmentation. In contrast, Faceness \cite{yang2015facial} needs additional hair, eyes, nose, mouth and beard annotations to train several attribute-aware face models and uses bounding box regression to refine detected windows. DPM and HeadHunter \cite{mathias2014face} use extra annotation to train view-specific components to tackle facial variations. JointCascade \cite{chen2014joint} uses face alignment to help face detection with manually labeled 27 facial points. Some qualitative results on PASCAL Face and FDDB dataset together with the face response heatmap are shown in Figure \ref{fig:detected_examples}.

In addition, we design several comparative experiments to investigate the impact of data augmentation. The CNN binary classifier in the second stage is retrained with a slightly different training dataset, while all the other components are left unchanged. For face images, we remove the darkened, blurred and occluded face images (Figure \ref{fig:training_example}(a)) respectively, and then padded and double-sized images (Figure \ref{fig:training_example}(b)) are removed from non-face images. As before, all the newly trained models are tested on FDDB dataset as shown in Figure \ref{fig:result_fddb}(e) and (f). We can see that the type of training dataset has a significant influence on the performance of the detector. For example, the padded and double-sized non-face images can help the model locate face area more precisely, i.e., neither too big nor too small. Therefore, better data augmentation makes full use of the high-capacity of CNN to achieve better performance. In our case, all kinds of facial variations can be represented in a single CNN model.

\section{Concluding Remarks and Future work}
In this paper, we have developed a method to exploit the internal representation power of hidden units of a deep learning neural network. We explicitly set out to seek convolutional channels that specifically respond to certain objects. Through a purposefully designed face specific training samples, we show that we can fine-tune a pretrained CNN to reinforce the internal face specific features. Use face detection as a case study, we have showed that object specific convolutional channels can be used to build face detectors, which has the advantage of being simple and still achieves state of the art performances. Our method can be extended to other objects, and we are currently applying it to objects from cars in photography images to cells in medical images.

%In this paper, we propose a robust face detector ContourFace for unconstrained settings based on convolutional neural networks. Our approach directly uses multi-resolution convolutional channel feature to extract face corresponding heatmap, which can be directly used for face detection or help select candidate face windows for refining detection. Additionally, we use a carefully designed dataset to train a discriminative classifier, which can capture all kinds of facial variations. We show that our detector can achieve similar state-of-the-art results without extra information such as facial landmarks and pose annotations. In the end, by using comparative experiments we demonstrate that the final performance can benefit from rich representative capacity of CNNs through well-designed data augmentation.

{\small
\bibliographystyle{ieee}
\bibliography{egbib}
}

\end{document}